\pdfoutput=1
\documentclass[11pt,letterpaper]{article}

\usepackage[margin=1in]{geometry}
\usepackage{amsmath,amssymb,amsthm,mathtools}
\usepackage[dvipsnames]{xcolor}
\usepackage{graphicx}
\usepackage[colorlinks=true,linkcolor=blue!60!black,citecolor=blue!60!black,urlcolor=blue!60!black,hypertexnames=false]{hyperref}
\usepackage[round]{natbib}
\usepackage[capitalize,nameinlink]{cleveref}
\usepackage{booktabs}
\usepackage{enumitem}
\hypersetup{pdftitle={Attention Can Hold Tokens Apart: Stable Clusters Beyond the Value Spectrum in Causal Self-Attention}}

\theoremstyle{plain}
\newtheorem{theorem}{Theorem}[section]
\newtheorem{proposition}[theorem]{Proposition}
\newtheorem{corollary}[theorem]{Corollary}
\newtheorem{lemma}[theorem]{Lemma}
\theoremstyle{definition}

\newtheorem{example}[theorem]{Example}
\theoremstyle{remark}
\newtheorem{remark}[theorem]{Remark}

\crefname{lemma}{Lemma}{Lemmas}
\crefname{theorem}{Theorem}{Theorems}
\crefname{proposition}{Proposition}{Propositions}
\crefname{corollary}{Corollary}{Corollaries}
\crefname{remark}{Remark}{Remarks}
\crefname{example}{Example}{Examples}

\newcommand{\R}{\mathbb{R}}
\newcommand{\Sph}{\mathbb{S}}
\newcommand{\T}{\top}
\newcommand{\ip}[2]{\langle #1,#2\rangle}
\newcommand{\diag}{\operatorname{diag}}
\newcommand{\Pp}{\mathbf{P}}
\DeclarePairedDelimiter{\abs}{\lvert}{\rvert}
\DeclarePairedDelimiter{\norm}{\lVert}{\rVert}

\title{Attention Can Hold Tokens Apart:\\ Stable Clusters Beyond the Value Spectrum\\ in Causal Self-Attention}
\author{Jinran Yi}
\date{}

\begin{document}
\maketitle

\begin{abstract}
Deep stacks of self-attention make token representations cluster, and the idealized dynamics used to study this
suggest a simple rule: the value matrix decides where tokens go. Under causal masking the first token follows the
leading eigenvector of the value matrix, and later tokens have been expected to collapse onto that direction or its
antipode. We show that the query--key geometry can overrule this rule. When the value matrix has distinct positive
eigenvalues, attention scores can hold tokens away from both poles: we construct configurations that attract an open
set of initial conditions to an off-eigenvector direction, including ones whose query--key product is positive
definite with invertible tied queries and keys, and a stable configuration of three clusters pairwise at least $90^\circ$ apart; these
attractors persist under a normalized residual update with any step size. For two tokens in the plane we derive the
exact phase law: with a diagonal score matrix $\diag(a,b)$, $a\ge 0$, an off-axis attractor exists for some value
spectrum if and only if $b-a>(1+a)/2$, inside a window of value ratios, so a score gap of $1/2$ is the sharp
uniform threshold. When the top eigenvalue is repeated, the picture reverses: every equilibrium other than consensus
is unstable, a token facing a consensus prefix joins it at an explicit exponential rate, and all tokens reach a single
cluster from almost every initialization when the sequence is no longer than the multiplicity or the complement is
one-dimensional. Where causally masked tokens cluster is thus decided jointly by the value spectrum and the query--key
geometry. As consequences, we disprove the two-cluster conjecture of Karagodin, Polyanskiy and Rigollet (2024), prove
their single-cluster conjecture in several regimes, and show that off-axis attractors appear for every score gap beyond
the constant used in recent positive results.
\end{abstract}


\section{Introduction}\label{sec:intro}

As a deep transformer processes a sequence, the representations of its tokens tend to become alike. Pure attention
loses rank doubly exponentially with depth \citep{dong2021attention}, and a line of mathematical work explains such
behaviour by viewing the tokens as particles on a sphere that interact through attention, with depth playing the role
of time \citep{lu2019understanding,sander2022sinkformers,geshkovski2023emergence,geshkovski2025mathematical}. In this
picture the particles form clusters, and the shape of the limit is governed by the value matrix: in the words of
\citet{geshkovski2023emergence}, ``the type of limiting object that emerges depends on the spectrum of the value
matrix.''

Generative language models use \emph{causal} attention, in which a token attends only to itself and to the tokens
before it. Causality makes the first token special: it attends to nothing but itself, so its trajectory is a
normalized power iteration of the value matrix $V$ and it converges to the leading eigenvector $\xi$ of $V$ or to its
antipode. The first token therefore acts as an anchor for everything that follows. A natural guess is that every later
token is eventually dragged to the same axis, so that tokens end in at most two clusters, $\xi$ and $-\xi$, placed by
the value spectrum alone. For $V=I$ and arbitrary queries and keys this is a theorem: all tokens merge into a single
cluster \citep{karagodin2024clustering}. For value matrices with distinct positive eigenvalues the two-cluster picture
was conjectured by \citet{karagodin2024clustering}, and \citet{nguyen2026two} recently proved two-cluster behaviour
under a block structure of the query--key product. This paper asks the question in its plain form:
\emph{does the value spectrum decide where causally masked tokens cluster, and how many clusters can survive?}

The answer depends on whether the top eigenvalue of $V$ is simple.

\paragraph{A simple top eigenvalue: attention can hold tokens apart.} A later token feels two pulls. Its predecessors,
which sit on the axis of $\xi$, pull it towards themselves; its own value dynamics push it towards the nearer pole.
The weights of the two pulls are set by attention, and the attention a token pays to itself depends on where the token
is. When the query--key geometry makes self-attention large away from the axis, the two pulls can balance stably at a
point that is neither $\xi$ nor $-\xi$. We make this precise.
\begin{itemize}[leftmargin=1.5em,itemsep=2pt]
\item \textbf{Stable off-eigenvector clusters} (\cref{thm:offaxis}). For $V=\diag(8,3)$ we give explicit score
matrices for which a two-token configuration with the second token off the axis of $\xi$ is a hyperbolic sink. Its
basin is open, so a positive fraction of all initializations ends there (about $38\%$ of uniformly random ones in our
simulations). The score matrix can be taken positive definite and realized with invertible tied queries and keys
$Q=K$; the construction extends to every dimension $d\ge 2$ and every sequence length $n\ge 2$; and the same
configuration stays attracting for the normalized residual update $x\mapsto (x+\eta h)/\norm{x+\eta h}$ at every
step size $\eta>0$.
\item \textbf{An exact phase law} (\cref{thm:phase,thm:threshold}). For two tokens in the plane with
$V=\diag(\lambda,\mu)$, $\lambda>\mu>0$, and score matrix $\diag(a,b)$, we classify all equilibria and their
stability through one scalar equation. For $a\ge 0$, an off-axis attractor exists for some value spectrum if and only
if $b-a>(1+a)/2$, and then exactly for the value ratios $(\lambda-\mu)/\lambda$ in an explicit open interval. The set
of score gaps $b-a$ for which some positive-definite diagonal score matrix admits an off-axis attractor is exactly
$(1/2,\infty)$; in the parametrization of \citet{nguyen2026two}, off-axis attractors appear for every gap
$m_1-m_0>1/(2\beta)$, just beyond their block condition $m_1<m_0+1/(2\beta)$. For an arbitrary planar score matrix, the second token has at most three attracting
destinations, and this is sharp (\cref{prop:five}).
\item \textbf{More than two clusters} (\cref{thm:three}). For $n=3$ tokens in the plane we give a hyperbolic sink at
which the three tokens point in three different directions, pairwise at least $90^\circ$ apart. Neither the
locations nor the number of clusters is dictated by the value spectrum.
\end{itemize}

\paragraph{A repeated top eigenvalue: consensus returns.} If the top eigenvalue $\lambda$ has an eigenspace $L$ of
dimension at least two (and the rest of the spectrum is dominated in the sense of \eqref{eq:H}), the mechanism above
breaks down, because a token trapped off the axis can now escape inside $L$.
\begin{itemize}[leftmargin=1.5em,itemsep=2pt]
\item \textbf{Every equilibrium other than consensus is unstable} (\cref{thm:unstable}), for every score matrix and
every sequence length, and the consensus set is normally attracting. Adding a repeated copy of the top eigenvalue to
the example of \cref{thm:offaxis} turns its sink into a saddle with expansion rate exactly $10/3$
(\cref{ex:bridge}).
\item \textbf{A token facing a consensus prefix joins it} (\cref{thm:frozen}), from every initial position outside a
null set, through a monotone coordinate that yields an explicit exponential rate.
\item \textbf{Almost-sure consensus in two regimes} (\cref{thm:fewtokens,thm:codim1}): when the number of tokens is
at most $\dim L$, and, for every number of tokens, when the complement of $L$ is one-dimensional.
\end{itemize}

Taken together, the results say that where causally masked tokens cluster is decided jointly by the value spectrum and
by the query--key geometry. A simple top eigenvalue leaves room for the attention scores to create their own stable
meeting points; a repeated top eigenvalue removes that room.

\paragraph{Relation to the conjectures of \citet{karagodin2024clustering}.} \Cref{thm:offaxis} disproves their
Conjecture~1, which predicts that for value matrices with distinct positive eigenvalues every token converges to $\pm\xi$
for almost every initialization. \Cref{thm:fewtokens,thm:codim1} prove their Conjecture~2 (single-cluster consensus
for a repeated top eigenvalue) in the stated regimes; the general case remains open. Our
examples do not contradict the theorems of \citet{nguyen2026two}: they all violate the block structure assumed there,
which our phase law shows to be essential (\cref{rem:nguyen}).

\paragraph{Organization.} \Cref{sec:setting} sets up the model and three elementary tools. \Cref{sec:offaxis} gives the
stable off-axis clusters, \cref{sec:phase} the planar phase law, \cref{sec:three} the three-cluster sink, and
\cref{sec:repeated} the consensus results for a repeated top eigenvalue. \Cref{sec:discussion} discusses scope
and related work. Proofs omitted from the main text are in the appendices, and a script that checks every
exact identity and every sign used in the examples is provided with the source files.

\section{Setting and three tools}\label{sec:setting}

We study $n$ tokens $x_1,\dots,x_n$ on the unit sphere $\Sph^{d-1}\subset\R^d$, $d\ge 2$, driven by fixed real
$d\times d$ matrices: a \emph{score matrix} $A=\beta Q^\T K$ (the inverse temperature $\beta>0$ and the query and key
matrices $Q,K$ enter only through $A$) and a \emph{value matrix} $V$. For $1\le j\le i\le n$ let
\begin{equation}\label{eq:weights}
p_{ij}(X)=\frac{e^{x_i^\T A x_j}}{\sum_{\ell\le i}e^{x_i^\T A x_\ell}},\qquad
h_i(X)=\sum_{j\le i}p_{ij}(X)\,Vx_j,\qquad \alpha_i(X)=\ip{x_i}{h_i(X)},
\end{equation}
and consider the causal self-attention flow
\begin{equation}\tag{CSA}\label{eq:csa}
\dot x_i=h_i-\ip{x_i}{h_i}\,x_i=\Pp_{x_i}h_i,\qquad i=1,\dots,n,
\end{equation}
where $\Pp_x=I-xx^\T$ projects onto the tangent space of the sphere at $x$. This is the dynamics (CSA) of
\citet{karagodin2024clustering} (their weights are $e^{\beta\ip{Qx_i}{Kx_j}}$, i.e.\ $e^{x_i^\T Ax_j}$); the self-edge
$j=i$ is included. The vector field is smooth on the compact manifold $(\Sph^{d-1})^n$, so \eqref{eq:csa} defines a
smooth flow, and ``almost every initialization'' refers to the product surface measure. Since $x_i^\T Ax_j$ ranges in
$[-\norm{A},\norm{A}]$, every weight satisfies
\begin{equation}\label{eq:weightlb}
p_{ij}(X)\ge e^{-2\norm{A}}/i .
\end{equation}
An equilibrium is a configuration with $h_i=\alpha_i x_i$ for all $i$; it is a \emph{hyperbolic sink} if every
eigenvalue of the Jacobian of \eqref{eq:csa} on the tangent space $\prod_i T_{x_i}\Sph^{d-1}$ has negative real part.
A hyperbolic sink attracts an open neighbourhood, so its basin has positive measure. Two symmetries will be used:
$X\mapsto -X$ maps solutions to solutions, and if $A$ and $V$ are diagonal so does every coordinate reflection applied to
all tokens.

\begin{lemma}[The first token]\label{lem:first}
$x_1(t)=e^{tV}x_1(0)/\norm{e^{tV}x_1(0)}$. If $V$ has a simple real top eigenvalue $\lambda$ with unit eigenvector $\xi$
and all other eigenvalues have real part below $\lambda$, then $x_1(t)\to\operatorname{sign}(\ip{\xi'}{x_1(0)})\,\xi$
exponentially fast for every $x_1(0)$ outside a hyperplane, where $\xi'$ is the corresponding left eigenvector. If
$V=\lambda I_L\oplus B$ with $\ip{Bz}{z}\le(\lambda-\delta)\norm z^2$ on $L^\perp$ (as in \eqref{eq:H} below), then
$x_1(t)\to\Pp_Lx_1(0)/\norm{\Pp_Lx_1(0)}$ at rate $e^{-\delta t}$ whenever $\Pp_Lx_1(0)\neq0$.
\end{lemma}

\begin{proof}
Token $1$ attends only to itself, so $h_1=Vx_1$ and \eqref{eq:csa} is the projection of the linear flow
$\dot y=Vy$ onto the sphere; normalizing $e^{tV}x_1(0)$ solves it. The first limit follows from the spectral
decomposition of $e^{tV}$. In the second case $e^{tV}x=e^{\lambda t}\Pp_Lx+e^{tB}\Pp_{L^\perp}x$ and
$\frac{d}{dt}\norm{e^{t(B-\lambda)}z}^2\le-2\delta\norm{e^{t(B-\lambda)}z}^2$, so $\norm{e^{t(B-\lambda)}}\le e^{-\delta t}$.
\end{proof}

\begin{lemma}[Causal triangularity and the diagonal blocks]\label{lem:jac}
The right-hand side of \eqref{eq:csa} for token $i$ depends only on $x_1,\dots,x_i$, so the Jacobian at any
configuration is block lower triangular and its spectrum is the union of the spectra of the diagonal blocks. At an
equilibrium $X^\ast$, let $U_i$ be a $d\times(d-1)$ matrix with orthonormal columns spanning $x_i^\perp$. The $i$-th
diagonal block is $J_i=U_i^\T(D_ih_i-\alpha_iI)U_i$, where
\begin{equation}\label{eq:block}
D_ih_i=p_{ii}V+\sum_{j\le i}p_{ij}\,(Vx_j-h_i)\,g_{ij}^\T,\qquad g_{ij}=Ax_j\ (j<i),\quad g_{ii}=(A+A^\T)x_i,
\end{equation}
is the derivative of $h_i$ with respect to $x_i$.
\end{lemma}

\begin{proof}
Triangularity is causality. The score $x_i^\T Ax_j$ has gradient $Ax_j$ in $x_i$ for $j<i$ and $(A+A^\T)x_i$ for
$j=i$; the softmax derivative is $\partial_{x_i}p_{ij}=p_{ij}(g_{ij}-\sum_\ell p_{i\ell}g_{i\ell})$. Hence
$D_ih_i=p_{ii}V+\sum_jVx_j\,\partial_{x_i}p_{ij}^\T$, which is \eqref{eq:block} because $\sum_jp_{ij}Vx_j=h_i$.
Differentiating $h_i-\ip{x_i}{h_i}x_i$ in a tangent direction $v\perp x_i$ at an equilibrium ($h_i=\alpha_ix_i$, so
$\ip{h_i}{v}=0$) gives $D_ih_i\,v-\alpha_iv$ plus a multiple of $x_i$, which $U_i^\T$ removes.
\end{proof}

\begin{lemma}[Normalized residual update]\label{lem:residual}
Let $\Psi_i(X)=(x_i+\eta h_i(X))/\norm{x_i+\eta h_i(X)}$ with $\eta>0$. At an equilibrium $X^\ast$ of \eqref{eq:csa}
with $1+\eta\alpha_i>0$ for all $i$, $\Psi$ fixes $X^\ast$, its derivative is block lower triangular, and its $i$-th
diagonal block is $I+\frac{\eta}{1+\eta\alpha_i}J_i$. In particular, if $J_i$ is $1\times1$ with value $-\gamma_i$ and
$0<\gamma_i\le 2\alpha_i$, the corresponding multiplier lies in $(-1,1)$ for every $\eta>0$.
\end{lemma}

\begin{proof}
At $X^\ast$, $x_i+\eta h_i=(1+\eta\alpha_i)x_i$, so $\Psi_i(X^\ast)=x_i$. The derivative of $z\mapsto z/\norm z$ at
$cx_i$, $c>0$, is $c^{-1}\Pp_{x_i}$, and $U_i^\T\Pp_{x_i}(I+\eta D_ih_i)U_i=I+\eta(J_i+\alpha_iI)$; dividing by
$c=1+\eta\alpha_i$ gives the block. The multiplier $1-\eta\gamma_i/(1+\eta\alpha_i)$ is below $1$, and it exceeds $-1$
because $\eta\gamma_i\le 2\eta\alpha_i<2(1+\eta\alpha_i)$.
\end{proof}

\section{Attention can hold a token off the top eigenvector}\label{sec:offaxis}

\begin{theorem}[Stable off-eigenvector clusters]\label{thm:offaxis}
Let $d=n=2$, $V=\diag(8,3)$, and let $A$ be any of
\[
A_0=\diag\!\Big(0,\tfrac{25}{9}\log 2\Big),\qquad
A_1=\diag\!\Big(\tfrac1{10},\tfrac{25}{9}\log2-\tfrac25\Big),\qquad
A_2=\diag\!\Big(\tfrac{\log 2}{9},\tfrac{7\log 2}{3}\Big).
\]
The configuration $X^\ast=(e_1,q)$, $q=(-\tfrac45,\tfrac35)$, is an equilibrium of \eqref{eq:csa} with attention
weights $(p_{21},p_{22})=(\tfrac13,\tfrac23)$ and $h_1=8e_1$, $h_2=2q$. Its Jacobian has the eigenvalues $-5$ and
\[
\nu_0=\tfrac65-\tfrac{64}{15}\log 2,\qquad \nu_1=\tfrac{258}{125}-\tfrac{64}{15}\log2,\qquad
\nu_2=\tfrac65-\tfrac{248}{75}\log 2
\]
for $A_0,A_1,A_2$ respectively, all negative; so $X^\ast$ is a hyperbolic sink, and every initialization in its open
basin has $x_2(t)\to q$, which is at distance $\sqrt{2/5}$ from $-e_1$ and $\sqrt{18/5}$ from $e_1$. Moreover:
\begin{enumerate}[label=(\roman*),leftmargin=2em,itemsep=1pt]
\item $A_1$ and $A_2$ are positive definite, so $A=Q^\T K$ with $Q=K=A^{1/2}$ invertible ($\beta=1$).
\item (All dimensions and lengths.) For $d\ge 3$, append to $V$ distinct eigenvalues $\nu_3,\dots,\nu_d\in(0,3)$ and to
$A$ arbitrary diagonal entries (positive ones keep $A_1,A_2$ positive definite). Then $(e_1,q)$, with $q$ embedded in the first two coordinates, remains a
hyperbolic sink of the two-token system. For every $n\ge 2$ the first two tokens do not depend on the others, so the
set of initializations of $n$ tokens with $x_2(t)\to q$ contains an open set.
\item (Discrete updates.) For every step size $\eta>0$, $X^\ast$ is an attracting fixed point of the normalized
residual update of \cref{lem:residual}.
\end{enumerate}
In particular, for every $d\ge2$ and $n\ge2$ there are value matrices with $d$ distinct positive eigenvalues, tied
invertible queries and keys, and an open set of initializations for which $x_2(t)$ converges to neither $\xi=e_1$ nor
$-\xi$.
\end{theorem}

\begin{proof}
\emph{Equilibrium.} For $A_0$, $q^\T A_0e_1=0$ and $q^\T A_0q=\frac{25}{9}\log2\cdot\frac{9}{25}=\log2$, so
$(p_{21},p_{22})=(\frac13,\frac23)$. For $A_1$ both scores are shifted by $-\frac{2}{25}$ ($q^\T A_1e_1=-\frac{2}{25}$
and $q^\T A_1q=\log2-\frac2{25}$) and for $A_2$ they are $-\frac{4\log2}{45}$ and $\frac{41\log2}{45}$; in each case
the difference is $\log 2$ and the weights are unchanged. Then
$h_2=\frac13(8,0)+\frac23(-\frac{32}5,\frac95)=(-\frac85,\frac65)=2q$ and $h_1=Ve_1=8e_1$.

\emph{Spectrum.} By \cref{lem:jac} the Jacobian is lower triangular in the angular coordinates. The first block is
$U_1^\T(V-8I)U_1=3-8=-5$. For the second, put $x_2=(\cos\theta,\sin\theta)$ with $x_1=e_1$ fixed; for
$A=\diag(a,b)$ the angle obeys (\cref{prop:angle} below)
\[
\dot\theta=-\sin\theta\,\frac{8e^{a\cos\theta}+5\cos\theta\,e^{a\cos^2\theta+b\sin^2\theta}}{e^{a\cos\theta}+e^{a\cos^2\theta+b\sin^2\theta}},
\]
and \cref{prop:angle} gives the derivative at $\cos\theta=-u$, $u=\frac45$, as
$\frac{5(1-u^2)}{1+ru}\,(1+au+2(a-b)u^2)$ with $r=\frac58$, $1+ru=\frac32$. Substituting $(a,b)$ gives $\nu_0,\nu_1,\nu_2$.
Their signs follow from $\log2>\frac12$: $\nu_0<\frac65-\frac{32}{15}<0$, $\nu_1<\frac{258}{125}-\frac{32}{15}=-\frac{26}{375}$ and
$\nu_2<\frac65-\frac{124}{75}<0$. A Hurwitz Jacobian makes $X^\ast$ a hyperbolic sink with an open basin
(e.g.\ by a quadratic Lyapunov function $y^\T Py$ with $J^\T P+PJ=-I$ in a chart). Finally
$\norm{q+e_1}^2=\frac1{25}+\frac9{25}=\frac25$ and $\norm{q-e_1}^2=\frac{81}{25}+\frac{9}{25}=\frac{18}5$.

\emph{(i)} The diagonal entries of $A_1$ and $A_2$ are positive because $\log2>\frac12$.

\emph{(ii)} At $X^\ast$ both tokens have zero components along $e_k$, $k\ge3$, and $A$ is diagonal, so the scores
have zero derivative in these directions and the softmax terms of \eqref{eq:block} vanish on them. The transverse
eigenvalues are therefore $\nu_k-8<0$ for token~1 and $p_{22}\nu_k-\alpha_2=\frac23(\nu_k-3)<0$ for token~2. The
first two tokens evolve autonomously, so their open basin times $(\Sph^{d-1})^{n-2}$ is open.

\emph{(iii)} Apply \cref{lem:residual}: $\alpha_1=8$ and $\gamma_1=5\le16$; $\alpha_2=2$ and, for all three matrices,
$\gamma_2=-\nu_\ell<\frac{64}{15}\log 2<\frac{64}{15}\cdot\frac34=\frac{16}5\le 4$.
\end{proof}

All identities and signs in \cref{thm:offaxis} are also verified in exact arithmetic by the accompanying script
(\texttt{checks.py}), which computes the Jacobian from the ambient vector field independently of \eqref{eq:block}.

\begin{remark}[How large is the basin?]\label{rem:basin}
The basin is not small. With $x_1$ held at $e_1$ and $x_2$ uniform on the circle, the two off-axis sinks attract the
fraction $\arccos(u_-)/\pi\approx0.4137$ of initial angles, where $u_-\approx0.2679$ is the unstable root of
\cref{prop:angle} (\cref{fig:phase}(c)). For the full two-token system with both tokens uniform, a numerical
integration of $20{,}000$ random initializations with $A_0$ ended at an off-axis configuration in $38.05\%$ of the runs
(all runs converged; a floating-point diagnostic, not part of any proof).
\end{remark}

\begin{remark}[Relation to \citet{nguyen2026two}]\label{rem:nguyen}
\citet{nguyen2026two} prove, for symmetric $V$ with a fixed eigenbasis, that a token whose predecessors converge to
$\pm\xi$ recurrently approaches $\pm\xi$ along a subsequence (outside one exceptional trajectory), under their
Assumption~4.1: $Q^\T K=\diag(m_0,\,m_1I+\mathcal B)$ with $\mathcal B$ skew, $m_0>0$ and $m_1<m_0+1/(2\beta)$. Their
abstract summarizes the result as two-cluster behaviour ``when $\lambda_{\max}>0$'', and their Remark~4.4 states that
the assumption covers the weight-tying setting of \citet{karagodin2024clustering}. All our examples have constant
(tied) parameters and a positive spectrum but lie outside Assumption~4.1: $A_0$ has $m_0=0$; $A_1$ and $A_2$ have
$\beta(m_1-m_0)=\frac{25}{9}\log2-\frac12\approx1.43$ and $\frac{20}{9}\log2\approx1.54$, both above $\frac12$; and the
score matrix of \cref{thm:three} is not block diagonal. So there is no contradiction; rather, two-cluster behaviour
needs the block condition and does not follow from $\lambda_{\max}>0$ and weight tying alone, and \cref{thm:threshold}
shows that off-axis attractors appear for every gap $m_1-m_0$ above the constant $1/(2\beta)$ of that condition. Off-axis limits do appear in their work in three
other forms, each different from \cref{thm:offaxis}: their Theorems~2.3 and~C.7 allow limits off $\pm e_1$ (in a
coordinate plane, under an error condition) without showing that they are attracting; their Theorem~4.3 exhibits an
open set of off-axis limits for the second token with the first token frozen at $e_1$ when both value eigenvalues are
\emph{negative}; and their Proposition~E.3 builds a single off-axis trajectory when an eigenvalue of $V$ varies in time.
\Cref{thm:offaxis} gives off-axis limits from an open set of initializations of the full system, with constant
parameters and a positive spectrum.
\end{remark}

\section{The planar phase law}\label{sec:phase}

Throughout this section $d=n=2$, $V=\diag(\lambda,\mu)$ with $\lambda>\mu>0$, and
\[
r=\frac{\lambda-\mu}{\lambda}\in(0,1).
\]
Write $x_2=(\cos\theta,\sin\theta)$. We first freeze $x_1=e_1$ (the \emph{anchored} flow), then return to the full
system.

The equilibrium equation \eqref{eq:F} below is the planar instance of the scalar condition $G(y_\ast)=0$ that
\citet[Thm.~C.7]{nguyen2026two} derive for limits off $\pm e_1$; what is new here is the stability analysis: which
roots attract, when attracting roots exist, and the resulting thresholds in the score and value parameters.

\begin{proposition}[Anchored angular equation]\label{prop:angle}
Let $A=\diag(a,b)$ and $x_1\equiv e_1$. Then
\begin{equation}\label{eq:theta}
\dot\theta=-\sin\theta\,\frac{\lambda w_1+(\lambda-\mu)\cos\theta\,w_2}{w_1+w_2},\qquad
w_1=e^{a\cos\theta},\quad w_2=e^{a\cos^2\theta+b\sin^2\theta}.
\end{equation}
Its zeros other than $\theta\in\{0,\pi\}$ are the points $x_2=(-u,\pm\sqrt{1-u^2})$, $0<u<1$, with
\begin{equation}\label{eq:F}
F(u):=\log(ru)+b+au+(a-b)u^2=0 .
\end{equation}
At such a zero the derivative of the right-hand side of \eqref{eq:theta} is
\begin{equation}\label{eq:nu}
\nu(u)=\frac{(\lambda-\mu)(1-u^2)}{1+ru}\,\big(1+au+2(a-b)u^2\big)=\frac{(\lambda-\mu)(1-u^2)}{1+ru}\,uF'(u),
\end{equation}
so the zero is attracting if $F'(u)<0$ and repelling if $F'(u)>0$. The point $\theta=0$ is always attracting
(derivative $-(2\lambda-\mu)/2$), and $\theta=\pi$ has derivative $\lambda(1-re^{2a})/(1+e^{2a})$, so $-e_1$ is
attracting if $S:=\log r+2a>0$ and repelling if $S<0$.
\end{proposition}

\begin{proof}
With $x_1=e_1$, the weights of token $2$ are proportional to $w_1=e^{x_2^\T Ae_1}$ and $w_2=e^{x_2^\T Ax_2}$, and
$\dot\theta=\ip{(-\sin\theta,\cos\theta)}{h_2}$ with $h_2=(w_1\lambda e_1+w_2Vx_2)/(w_1+w_2)$, which gives
\eqref{eq:theta}. Writing $u=-\cos\theta$, the bracket equals $\lambda w_1(1-ru\,w_2/w_1)$ and
$w_2/w_1=e^{b+au+(a-b)u^2}$, so $\dot\theta=-\lambda p_{21}\sin\theta\,(1-G(u))$ with $G=e^{F}$ and
$p_{21}=w_1/(w_1+w_2)$. A zero with $\sin\theta\ne0$ needs $G(u)=1$, which forces $u>0$, i.e.\ \eqref{eq:F}. At such a
zero, $d\dot\theta/d\theta=\lambda p_{21}\sin^2\theta\,F'(u)$ (as $du/d\theta=\sin\theta$), and $p_{21}=ru/(1+ru)$
there because $w_2/w_1=1/(ru)$; this is \eqref{eq:nu}, using $uF'(u)=1+au+2(a-b)u^2$. The derivatives at $\theta=0$
and $\theta=\pi$ follow by linearizing \eqref{eq:theta}.
\end{proof}

\begin{theorem}[Classification for $a\ge0$]\label{thm:phase}
Let $A=\diag(a,b)$ with $a\ge 0$, $S=\log r+2a$, and $M_F=\sup_{0<u\le1}F(u)$. Away from the critical cases $S=0$
and $M_F=0$, the anchored flow \eqref{eq:theta} has exactly the following equilibria and attractors:
\begin{center}
\begin{tabular}{lll}
\toprule
regime & off-axis equilibria & attracting equilibria of token 2\\
\midrule
$S>0$ & one symmetric pair, repelling & $e_1$ and $-e_1$\\
$S<0$, $M_F<0$ & none & $e_1$ only\\
$S<0$, $M_F>0$ & two symmetric pairs, at $u_-<u_+$ & $e_1$ and the pair at $u_+$\\
\bottomrule
\end{tabular}
\end{center}
Every trajectory of the anchored flow converges to one of its equilibria. In the full two-token system, each
attracting equilibrium $(e_1,x_2^\ast)$ listed above is a hyperbolic sink.
\end{theorem}

\begin{proof}
$F(0^+)=-\infty$, $F(1)=S$, $F'(u)=1/u+a+2(a-b)u$ and $F''(u)=-1/u^2+2(a-b)$. If $b\le a$ then $F'>0$ on $(0,1]$.
If $b>a$ then $F$ is strictly concave, and either $F'(1)\ge0$, so again $F'>0$ on $(0,1)$, or $F$ has a unique
interior maximizer $u_\ast$ with $M_F=F(u_\ast)>S$. If $F$ is increasing it has one zero, with $F'>0$, when $S>0$ and
none when $S<0$. If $F$ has an interior maximum and $S>0$, it has exactly one zero, on the increasing branch; if
$S<0$ it has two zeros $u_-<u_\ast<u_+$ when $M_F>0$ and none when $M_F<0$, with $F'(u_-)>0>F'(u_+)$. The signs of
$\nu$ and of the pole derivatives come from \cref{prop:angle}. On the circle, a trajectory of an autonomous scalar
equation with finitely many zeros is monotone between consecutive zeros and converges to one of them. In the full
system, token $1$ has eigenvalue $\mu-\lambda<0$ at $e_1$, so by \cref{lem:jac} an anchored sink with nonzero
derivative is a hyperbolic sink of the two-token system.
\end{proof}

\begin{theorem}[Exact threshold and sharp score gap]\label{thm:threshold}
Let $A=\diag(a,b)$ with $a\ge 0$. There exist $\lambda>\mu>0$ for which the two-token system has an off-axis
attracting equilibrium (equivalently, a hyperbolic off-axis sink) if and only if
\begin{equation}\label{eq:threshold}
b-a>\frac{1+a}{2}.
\end{equation}
In that case the admissible value ratios form the open interval $e^{-M_0}<r<e^{-2a}$, where
$M_0=\max_{0<u\le1}\{\log u+b+au+(a-b)u^2\}>2a$. Consequently, the set of score gaps $b-a$ for which some
positive-definite diagonal $A$ (with some $V$) has an off-axis attractor is exactly $(1/2,\infty)$, while no diagonal
$A$ with $a\ge0$ and $b-a\le 1/2$ has one for any $V=\diag(\lambda,\mu)$, $\lambda>\mu>0$.
\end{theorem}

\begin{proof}
An attracting off-axis equilibrium of the full system has $x_1\in\{\pm e_1\}$: at an equilibrium $x_1$ is an
eigenvector of $V$, and at $x_1=\pm e_2$ token $1$ has the expanding eigenvalue $\lambda-\mu$. By the symmetry
$X\mapsto-X$ we may take $x_1=e_1$, and then $x_2$ must be an attracting off-axis zero of \eqref{eq:theta}. By the
proof of \cref{thm:phase}, an off-axis zero with $F'\le0$ exists if and only if $F$ has an interior maximum, $S<0$ and
$M_F\ge0$ (if $F$ is increasing on $(0,1]$, every zero has $F'>0$; at $M_F=0$ the zero is a saddle-node, attracting
from one side only, and we do not count it). An interior maximum exists if and only if $b>a$ and $F'(1)<0$, i.e.\
$1+a+2(a-b)<0$, which is \eqref{eq:threshold}. Writing $F=\log r+H$ with $H(u)=\log u+b+au+(a-b)u^2$, the
conditions $S<0<M_F$ read $\log r<-2a$ and $\log r>-M_0$; the interval is nonempty because $M_0=\max H>H(1)=2a$ when
the maximum is interior, and it lies in $(0,1)$ because $a\ge0$. Every $r\in(0,1)$ is realized by
$\mu=\lambda(1-r)$. For the last statement: if $b-a=g>1/2$, choose $0<a<2g-1$; then \eqref{eq:threshold} holds and
$A=\diag(a,a+g)$ is positive definite. If $g\le 1/2$ then $g\le(1+a)/2$ for every $a\ge0$.
\end{proof}

In the parametrization $A=\beta\,\diag(m_0,m_1)$ of \citet{nguyen2026two}, condition \eqref{eq:threshold} reads
$m_1-m_0>\frac{m_0}{2}+\frac1{2\beta}$. Its infimum over $m_0>0$ is exactly the constant $\frac1{2\beta}$ of their gap
condition $m_1<m_0+\frac1{2\beta}$: for every gap $m_1-m_0>\frac1{2\beta}$ there are positive-definite planar examples with
off-axis attractors, while for a fixed $m_0>0$ the exact threshold is larger.

\begin{corollary}[The slice $a=0$]\label{cor:slice}
For $A=\diag(0,k)$ an off-axis hyperbolic sink exists if and only if $k>1/2$ and $r\,e^{k-1/2}>\sqrt{2k}$. For
$V=\diag(8,3)$ ($r=5/8$) the critical value is $k_c\approx1.5288$, and $A_0$ of \cref{thm:offaxis} has
$k=\frac{25}{9}\log 2\approx1.9254>k_c$. On the boundary $r\,e^{k-1/2}=\sqrt{2k}$ with $k>1/2$, the off-axis
equilibrium is a saddle-node that attracts a one-sided neighbourhood in the anchored flow.
\end{corollary}

\begin{proof}
With $a=0$, $H(u)=\log u+k(1-u^2)$ has its maximum at $u_\ast=1/\sqrt{2k}$ when $k>1/2$, with
$M_0=k-\frac12-\frac12\log(2k)$; the condition $r>e^{-M_0}$ is the stated inequality, and $S=\log r<0$ always. At
equality $F(u_\ast)=F'(u_\ast)=0>F''(u_\ast)$, so $\dot\theta$ has a double zero there and keeps one sign on both sides.
\end{proof}

\begin{figure}[t]
\centering
\includegraphics[width=\linewidth]{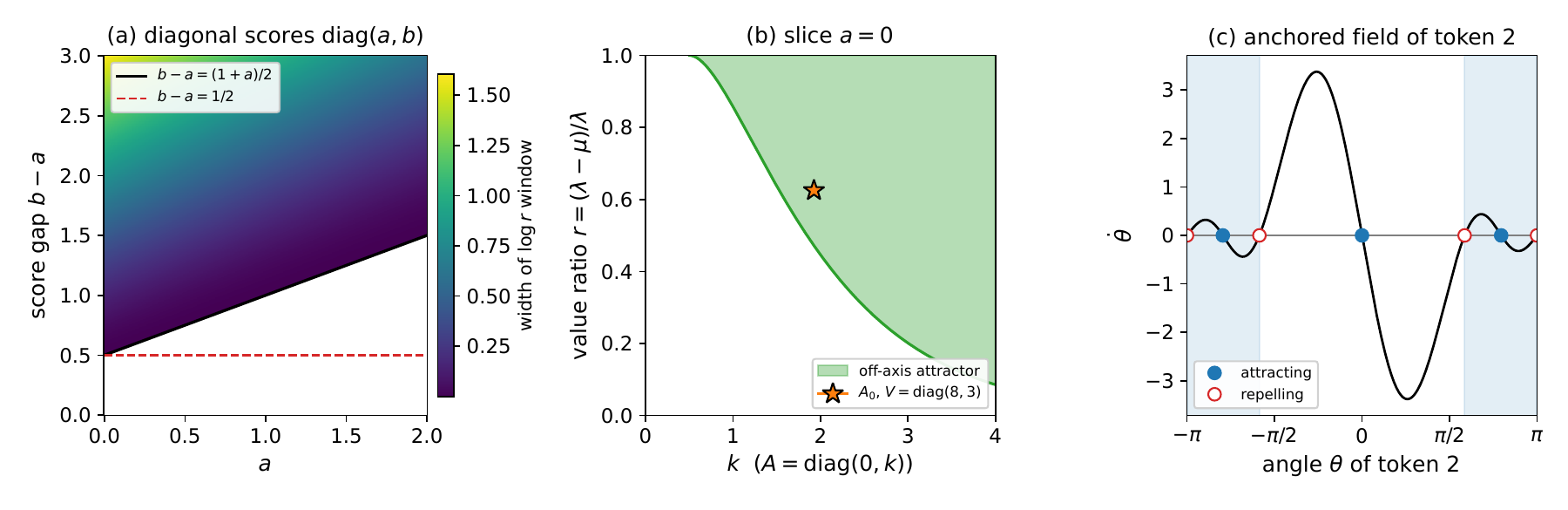}
\caption{The planar phase law. (a) Diagonal score matrices $A=\diag(a,b)$: off-axis attractors exist for some value
spectrum exactly above the line $b-a=(1+a)/2$ (\cref{thm:threshold}); colour shows the width $M_0-2a$ of the admissible
window of $\log r$. The dashed line $b-a=1/2$ is the uniform threshold. (b) The slice $a=0$ in the $(k,r)$ plane
(\cref{cor:slice}); the star is $A_0$, $V=\diag(8,3)$. (c) The anchored angular field of token~2 for this example:
attracting zeros at $\theta=0$ and at $x_2=(-\frac45,\pm\frac35)$, repelling zeros at $\theta=\pi$ and at
$u=u_-\approx0.268$; the shaded arcs are the basins of the off-axis attractors.}
\label{fig:phase}
\end{figure}

\begin{remark}[Reading the phase law]\label{rem:phase}
The threshold has a simple meaning. In \eqref{eq:F}, $b+au+(a-b)u^2$ is the score advantage of self-attention over
attention to the anchor at $x_2=(-u,\pm\sqrt{1-u^2})$. An off-axis attractor needs this advantage to \emph{decrease}
fast enough as $x_2$ moves towards the antipode $-e_1$ (i.e.\ as $u\to1$), so that a token pushed towards the pole
by its own value dynamics loses self-attention and is pulled back by the anchor, while a token moving towards the
equator gains self-attention and is pushed back towards the pole. Condition \eqref{eq:threshold} is exactly
$F'(1)<0$. The value ratio $r$ then only has to place the balance point inside the window. \Cref{fig:phase} shows the
phase diagram.
\end{remark}

The anchored picture describes the full two-token system for almost every initialization, as long as the anchored
equilibria are hyperbolic; this holds for every score matrix, not only diagonal ones.

\begin{proposition}[From the anchored flow to the full system]\label{prop:full}
Let $V=\diag(\lambda,\mu)$, $\lambda>\mu>0$, and let $A$ be any real $2\times2$ matrix. Suppose every zero of the anchored
angular field of $x_2$ for $x_1\equiv e_1$ is hyperbolic (by the symmetry $X\mapsto-X$ the same then holds for
$x_1\equiv-e_1$). Then for almost every
initialization of the two-token system, $x_1(t)\to\sigma e_1$ for some $\sigma\in\{\pm1\}$ and $x_2(t)$ converges
exponentially to an attracting zero of the anchored field for $x_1\equiv\sigma e_1$.
\end{proposition}

The proof (\cref{app:full}) combines the exponential convergence of $x_1$ (\cref{lem:first}) with the fact that a
scalar equation whose right-hand side converges exponentially in $C^1$ to a field with hyperbolic zeros can converge
to a repelling zero only from finitely many initial angles.

How many destinations can the second token have? For a general planar score matrix the answer is at most three,
and this is attained.

\begin{proposition}[At most three attracting destinations]\label{prop:five}
Let $V=\diag(\lambda,\mu)$, $\lambda>\mu>0$, $x_1\equiv e_1$, and $A=\begin{pmatrix}a_{11}&a_{12}\\a_{21}&a_{22}\end{pmatrix}$
arbitrary. The anchored flow of $x_2$ has at most $4$ equilibria other than $\pm e_1$, hence at most $6$ in total; if all
are hyperbolic, at most $3$ of them are attracting. All three bounds are attained, already by diagonal score matrices in
the regime $S<0<M_F$ of \cref{thm:phase}.
\end{proposition}

\begin{proof}
Parametrize the open half-circle $\{\cos\theta<0\}$, which contains every equilibrium other than $\pm e_1$
(\cref{prop:angle} applies verbatim to the bracket of the angular equation), by
$x_2=\big(-\frac{1-t^2}{1+t^2},\frac{2t}{1+t^2}\big)$, $-1<t<1$, so that $u=-\cos\theta=\frac{1-t^2}{1+t^2}$. The equilibrium
condition $\log(ru)+x_2^\T A(x_2-e_1)=0$, multiplied by $(1+t^2)^2$, reads $C(t)=W(t)$ with the cubic polynomial
\[
C(t)=2a_{11}-(2a_{12}+4a_{21})t+(4a_{22}-2a_{11})t^2+2a_{12}t^3
\]
and
\[
W(t)=(1+t^2)^2\Big(\ell+\log\frac{1+t^2}{1-t^2}\Big)=\ell+(2\ell+2)t^2+(\ell+4)t^4+\tfrac83t^6+\cdots,
\]
where $\ell=\log\frac1r>0$; this is an even power series with nonnegative coefficients (since $\log\frac{1+t^2}{1-t^2}=2\sum_{k\ \mathrm{odd}}t^{2k}/k$). Hence
$W^{(4)}(t)\ge24(\ell+4)>0$ on $(-1,1)$, while $C^{(4)}\equiv0$, and by Rolle's theorem $C-W$ has at most four zeros counted
with multiplicity. With the poles there are at most six equilibria; hyperbolic zeros of a vector field on the circle
alternate between attracting and repelling, so at most three attract. In the regime $S<0<M_F$ of \cref{thm:phase} there
are four off-axis equilibria and three attractors ($e_1$ and the pair at $u_+$).
\end{proof}

\citet{nguyen2026two} show that limits off $\pm e_1$ form a finite set (their Theorem~C.7); \cref{prop:five} gives the
sharp count in the plane.

\section{More than two clusters}\label{sec:three}

The attractor of \cref{thm:offaxis} has two clusters, but not at the poles. With three tokens, the scores can hold
three clusters at once.

\begin{theorem}[Three clusters]\label{thm:three}
Let $d=2$, $n=3$, $V=\diag(8,3)$, $L=\log 2$, $M=\log\frac{97}{48}$, and
\[
A=\begin{pmatrix}
-\frac89+\frac{14}{27}L+\frac5{12}M & -\frac83+\frac{20}{9}L+\frac54M\\[3pt]
\frac23-\frac{11}{9}L & 2
\end{pmatrix}.
\]
The configuration
\[
x_1=(1,0),\qquad x_2=\big(-\tfrac45,\tfrac35\big),\qquad x_3=\big(-\tfrac35,-\tfrac45\big)
\]
is a hyperbolic sink of \eqref{eq:csa}, with attention weights $(p_{21},p_{22})=\frac13(1,2)$ and
$(p_{31},p_{32},p_{33})=\frac1{157}(48,12,97)$, and $h_1=8x_1$, $h_2=2x_2$, $h_3=\frac{264}{157}x_3$. The Jacobian
eigenvalues are
\[
-5,\qquad -\tfrac{62}{15}+\tfrac{8}{45}L+\tfrac85M\approx-2.8845,\qquad
\tfrac{1687}{785}-\tfrac{3104}{471}+\tfrac{23524}{7065}L+\tfrac{1261}{785}M\approx-1.0032 .
\]
Hence an open set of initializations converges to three distinct directions, pairwise at least $90^\circ$ apart,
although $V$ has two distinct positive eigenvalues. The configuration is also an attracting fixed point of the
normalized residual update for every $\eta>0$.
\end{theorem}

\begin{proof}
Direct multiplication gives the score differences $x_2^\T A(x_2-x_1)=L$, $x_3^\T A(x_2-x_1)=-2L$ and
$x_3^\T A(x_3-x_1)=M$, hence the weights: row~2 is proportional to $(1,2)$ and row~3 to
$(1,\frac14,\frac{97}{48})$, i.e.\ to $(48,12,97)$. Then $h_2=2x_2$ as in \cref{thm:offaxis}, and
$h_3=V(48x_1+12x_2+97x_3)/157=\frac1{157}V(-\frac{99}5,-\frac{352}5)=\frac{264}{157}x_3$. The Jacobian is lower
triangular (\cref{lem:jac}) with $1\times1$ diagonal blocks; evaluating \eqref{eq:block} gives the three eigenvalues.
Their signs follow from $\frac{69}{100}<L<\frac{7}{10}$ and $\frac{7}{10}<M<\frac{71}{100}$: the second eigenvalue is
at most $-\frac{62}{15}+\frac{8}{45}\cdot\frac7{10}+\frac85\cdot\frac{71}{100}<-2.87$ and the third at most
$\frac{1687}{785}-\frac{3104}{471}+\frac{23524}{7065}\cdot\frac7{10}+\frac{1261}{785}\cdot\frac{71}{100}<-0.96$. The pairwise
inner products of $x_1,x_2,x_3$ are $-\frac45,-\frac35,0$. For the residual update use \cref{lem:residual} with
$\alpha=(8,2,\frac{264}{157})$; the same logarithm bounds give $\gamma_2<2.9<4$ and $\gamma_3<1.02<\frac{528}{157}$.
\end{proof}

The script \texttt{checks.py} re-derives the equilibrium and the full Jacobian of \cref{thm:three} from the ambient
vector field and certifies all three signs with rational bounds on the logarithms. In simulations from $20{,}000$ uniformly
random initial conditions, $2.0\%$ of runs converged to this configuration or its image under $X\mapsto -X$, and $20.0\%$
ended in three distinct directions, most of them at a second three-cluster equilibrium with $x_2=(-\frac45,\frac35)$ and
$x_3\approx(-0.923,0.385)$; all other runs ended in a single cluster (floating-point diagnostics, not used in any proof).

\section{A repeated top eigenvalue restores consensus}\label{sec:repeated}

We now assume that the top eigenvalue is repeated:
\begin{equation}\label{eq:H}\tag{H}
\R^d=L\oplus L^\perp,\quad m:=\dim L\ge2,\quad V=\lambda I_L\oplus B,\quad \lambda>0,\quad
\ip{Bz}{z}\le(\lambda-\delta)\norm z^2\ (z\in L^\perp)
\end{equation}
for some $\delta>0$, where $B:L^\perp\to L^\perp$ is an arbitrary real operator (it may be non-normal and have complex
or negative eigenvalues). This is the setting of Conjecture~2 of \citet{karagodin2024clustering} (their strict
inequality for $z\neq0$ gives a uniform $\delta$ by compactness). By \cref{lem:first}, for almost every $x_1(0)$,
\[
x_1(t)\to\xi:=\frac{\Pp_Lx_1(0)}{\norm{\Pp_Lx_1(0)}},
\]
where $\Pp_L$ is the orthogonal projection onto $L$. The conjecture is that $x_i(t)\to\xi$ for every $i$ and almost
every initialization. Throughout, for a token $x_i$ and a unit vector $\xi\in L$ we write
\begin{equation}\label{eq:decomp}
x_i=c_i\xi+z_i+y_i,\qquad z_i\in W:=L\cap\xi^\perp,\quad y_i\in L^\perp .
\end{equation}
The subspace $W$ has dimension $m-1\ge1$; it is the escape route that a simple top eigenvalue lacks.

\begin{example}[The trap of \cref{thm:offaxis} becomes a saddle]\label{ex:bridge}
Take $d=3$, $V=\diag(8,3,8)$ (so $L=\operatorname{span}(e_1,e_3)$, $B=3$) and $A=\diag(0,\frac{25}9\log2,0)$. The
configuration $(e_1,q)$, $q=(-\frac45,\frac35,0)$, is still an equilibrium, but its Jacobian eigenvalues are
$\{-5,\ 0,\ \frac65-\frac{64}{15}\log2,\ \frac{10}3\}$. The new direction $e_3$ is neutral for token~1 (it rotates the
consensus direction inside $L$) and expanding for token~2, with rate $p_{22}\cdot8-\alpha_2=\frac{16}3-2=\frac{10}3$.
More generally, appending an eigen-direction with value $\nu$ to $\diag(8,3)$ gives token~2 the eigenvalue
$\frac23(\nu-3)$: the trap survives exactly when $\nu$ is below the second eigenvalue $3$, and a repeated top
eigenvalue $\nu=8$ destroys it.
\end{example}

\begin{theorem}[Only consensus is stable]\label{thm:unstable}
Assume \eqref{eq:H}. For every $n$ and every score matrix $A$, every equilibrium of \eqref{eq:csa} outside the consensus
set $\mathcal C=\{(\xi,\dots,\xi):\xi\in L,\ \norm\xi=1\}$ has a Jacobian eigenvalue with positive real part. At every
point of $\mathcal C$ the Jacobian has the eigenvalue $0$ with multiplicity $m-1=\dim\mathcal C$ (directions tangent to
$\mathcal C$), and all other eigenvalues have negative real part: $-\lambda(1-\frac1i)$ on $W$ and the eigenvalues of
$\frac1iB-\lambda I$ on $L^\perp$ for token $i\ge2$, and those of $B-\lambda I$ for token $1$.
\end{theorem}

\begin{proof}
At an equilibrium, $Vx_1=\alpha_1x_1$, so $x_1$ is a real eigenvector of $V$. Under \eqref{eq:H} such a vector lies
either in $L$ or in $L^\perp$ (with an eigenvalue $\nu\le\lambda-\delta$ of $B$): indeed $Vx=\nu x$ forces
$(\lambda-\nu)\Pp_Lx=0$ and $B\Pp_{L^\perp}x=\nu\Pp_{L^\perp}x$, and $\nu=\lambda$ is impossible on $L^\perp\setminus\{0\}$.
If $x_1\in L^\perp$, then $L\subset x_1^\perp$ is invariant under $V-\nu I$, so the first block has the eigenvalue
$\lambda-\nu>0$.

Otherwise $x_1=\xi\in L$. Let $i$ be the first index with $x_i\ne\xi$ and use \eqref{eq:decomp} for $x=x_i$; the
predecessors all equal $\xi$, so with $p=\sum_{j<i}p_{ij}$ and $q=p_{ii}=1-p$,
\[
h_i=(p\lambda+q\lambda c)\,\xi+q\lambda\, z+qBy .
\]
Equilibrium $h_i=\alpha_ix_i$ gives $p\lambda+q\lambda c=\alpha_ic$, $q\lambda z=\alpha_iz$ and $qBy=\alpha_iy$. If
$z\neq0$ then $\alpha_i=q\lambda$ and the first relation gives $p\lambda=0$, impossible; so $z=0$. Consider the
tangent subspaces $W$ and $W'=x_i^\perp\cap(\operatorname{span}(\xi)\oplus L^\perp)$, so that
$T_{x_i}\Sph^{d-1}=W\oplus W'$. In \eqref{eq:block}, every vector $Vx_j-h_i$ lies in
$\operatorname{span}(\xi)\oplus L^\perp$, and $V$ maps $W$ to itself (as $\lambda I$) and $\operatorname{span}(\xi)\oplus
L^\perp$ to itself. Hence $J_i$ maps $W'$ into $W'$, and for $w\in W$, $J_iw=(q\lambda-\alpha_i)w+w'$ with $w'\in W'$.
In a basis adapted to $W\oplus W'$, $J_i$ is block triangular with the block $(q\lambda-\alpha_i)I_W$, so
$q\lambda-\alpha_i$ is an eigenvalue. If $y\ne0$, then $\alpha_i/q=\nu$ is a real eigenvalue of $B$ and
$q\lambda-\alpha_i=q(\lambda-\nu)\ge q\delta>0$. If $y=0$, then $x_i=-\xi$, $\alpha_i=\lambda(q-p)$ and
$q\lambda-\alpha_i=p\lambda>0$. By \cref{lem:jac} the full Jacobian has this positive eigenvalue.

At a consensus point all scores in a row coincide, so $p_{ij}=1/i$, $h_i=\lambda\xi$, $\alpha_i=\lambda$, and every
vector $Vx_j-h_i$ in \eqref{eq:block} vanishes. Hence $J_i=U_i^\T(\frac1iV-\lambda I)U_i$, which acts as
$\lambda(\frac1i-1)$ on $W$ and as $\frac1iB-\lambda I$ on $L^\perp$, whose symmetric part is at most
$(\frac{\lambda-\delta}{i}-\lambda)I\prec0$. For $i=1$ the $W$-block is $0$, giving the $m-1$ neutral directions.
\end{proof}

The next result shows directly why traps cannot form: a token whose predecessors have already agreed is always captured.

\begin{theorem}[A consensus prefix captures the next token]\label{thm:frozen}
Assume \eqref{eq:H}, fix $i\ge2$ and a unit vector $\xi\in L$, hold $x_1=\dots=x_{i-1}=\xi$, and let $x=x_i$ evolve by
\eqref{eq:csa}. Write $x=c\xi+z+y$ as in \eqref{eq:decomp}, let $p(x)=\sum_{j<i}p_{ij}$ and $q=1-p$, and
$D(y)=\lambda\norm y^2-\ip{By}{y}\ge\delta\norm{y}^2$. Away from $x=\xi$ the vector $w=z/(1-c)$ satisfies
\begin{equation}\label{eq:wmono}
\dot w=\Big(p\lambda+\frac{q\,D(y)}{1-c}\Big)w .
\end{equation}
Consequently, if $z(0)\neq0$ then
\begin{equation}\label{eq:frozenrate}
\norm{x(t)-\xi}\le\frac{2}{\norm{w(0)}}\,e^{-p_{\min}\lambda t},\qquad
p_{\min}=\frac{i-1}{i-1+e^{2\norm A}} .
\end{equation}
The exceptional set $\{z(0)=0\}$ is a great subsphere of codimension $m-1\ge1$, hence null.
\end{theorem}

\begin{proof}
Here $h=p\lambda\xi+qVx=(p\lambda+q\lambda c)\xi+q\lambda z+qBy$ and $\alpha=\ip xh=p\lambda c+q\lambda-qD(y)$, using
$c^2+\norm z^2=1-\norm y^2$. The equations $\dot c=p\lambda+q\lambda c-\alpha c$ and $\dot z=(q\lambda-\alpha)z$ give
\[
\frac{\dot w}{w}=q\lambda-\alpha+\frac{\dot c}{1-c}=\frac{(q\lambda-\alpha)(1-c)+p\lambda+q\lambda c-\alpha c}{1-c}
=\frac{\lambda-\alpha}{1-c},
\]
and $\lambda-\alpha=p\lambda(1-c)+qD(y)$, which is \eqref{eq:wmono}. Since $p\ge p_{\min}$ by the argument of
\eqref{eq:weightlb} (the predecessors carry total weight $(i-1)e^{s}/((i-1)e^s+e^{s'})$ with $s'-s\le2\norm A$),
$\norm{w(t)}\ge e^{p_{\min}\lambda t}\norm{w(0)}$. Finally $\norm z^2\le1-c^2$ gives $\norm w^2\le(1+c)/(1-c)\le
2/(1-c)$, and $\norm{x-\xi}^2=2(1-c)\le4/\norm w^2$.
\end{proof}

If $m=1$ (a simple top eigenvalue), $W=\{0\}$, $z\equiv0$, and \eqref{eq:wmono} is empty; this is exactly the room that
\cref{thm:offaxis} exploits. The frozen prefix is an idealization of the moving prefix; the next two theorems handle
moving prefixes in two regimes.

\begin{theorem}[Few tokens]\label{thm:fewtokens}
Assume \eqref{eq:H}. If there is $h\in L$ with $\ip{h}{x_i(0)}>0$ for all $i\le n$, then $x_i(t)\to\xi$ for all $i$. In
particular, if $n\le\dim L$, then $x_i(t)\to\xi$ for all $i$ and almost every initialization, for every score matrix
$A$ and every $B$ as in \eqref{eq:H}.
\end{theorem}

\begin{theorem}[One-dimensional complement]\label{thm:codim1}
Let $d\ge3$, let $e$ be a unit vector, $L=e^\perp$, and $V=\lambda\Pp_L+\mu ee^\T$ with $\mu<\lambda$, $\lambda>0$
($\mu$ may be negative). For every $n$ and every score matrix $A$, almost every initialization satisfies $x_i(t)\to\xi$
for all $i$.
\end{theorem}

Both proofs (\cref{app:repeated}) go by induction over the tokens. The key step is a capture lemma: once a token has a
nonnegative component along $\xi$ at a late time, dividing its other components by that component turns
\eqref{eq:csa} into a contraction (\cref{lem:capture}). For \cref{thm:fewtokens}, the half-space condition is preserved
by the flow and forces capture. For \cref{thm:codim1}, a token that is never captured must keep a negative component
along $\xi$ bounded away from zero; its $W$-component then dies out, its motion reduces to an asymptotically
autonomous equation on a circle, which converges to an equilibrium; every such equilibrium has an expanding direction in
$W$ (as in \cref{thm:unstable}), and a countable cover by local center-stable manifolds shows that the set of
initializations converging to these equilibria is null. Once convergence holds, it is exponential:

\begin{proposition}[Rate]\label{prop:rate}
Under \eqref{eq:H}, if a trajectory converges to $(\xi,\dots,\xi)$, then for every $0<\rho<\min(\delta,\lambda/2)$ there
is $C_\rho<\infty$, depending on the trajectory, with $\max_i\norm{x_i(t)-\xi}\le C_\rho e^{-\rho t}$.
\end{proposition}

The exponent $\lambda/2$ cannot be improved in general: for $A=0$, $x_1\equiv\xi$, and $x_2$ in a plane of $L$ through
$\xi$, the angle of $x_2$ obeys $\dot\theta=-\frac\lambda2\sin\theta$ (both weights equal $\frac12$); and $\delta$ cannot be
improved either, by taking $B=(\lambda-\delta)I$.

\section{Discussion}\label{sec:discussion}

\paragraph{What the results do and do not say.} All statements concern the idealized dynamics \eqref{eq:csa}: one
attention head with fixed, depth-shared matrices, no feed-forward layers, no positional encoding, and tokens normalized
to the sphere. Trained transformers do not satisfy these hypotheses literally; for instance, a single head's value map
$W_OW_V$ has rank at most the head dimension, and rotary encodings make the scores position dependent. The results
therefore do not predict how many clusters a given language model forms. What they show is a structural fact about the
model class that is used to reason about clustering: the value spectrum alone does not determine the limiting geometry,
and the attention scores can stabilize configurations that the value spectrum would not suggest. \Cref{thm:offaxis,thm:three}
are also not artefacts of continuous time, since the same configurations attract the normalized residual update at every
step size.

\paragraph{Related work.} The interacting-particle view of transformers goes back to \citet{lu2019understanding} and
\citet{sander2022sinkformers}; clustering theorems for full (non-causal) attention were proved by
\citet{geshkovski2023emergence,geshkovski2025mathematical}, and \citet{rigollet2025mean} surveys the mean-field theory.
For causal attention, \citet{karagodin2024clustering} proved single-cluster convergence for $V=I$ and arbitrary $Q,K$
and stated the two conjectures addressed here; \citet{nguyen2026two} proved two-cluster behaviour under the block
structure discussed in \cref{rem:nguyen}. \citet{abella2025consensus} prove consensus results for symmetric value
matrices with a simple top eigenvalue under initial hemisphere conditions, including a causal version proved by
induction over predecessors; the hemisphere argument behind \cref{thm:fewtokens} is of the same type, applied to a
repeated top eigenvalue, a direction selected by the first token, and a non-normal complement.
\citet{chen2025quantitative} give a partial resolution of Conjecture~2 adapted to non-causal mean-field dynamics.
Multistability of full (non-causal) self-attention with a simple positive top eigenvalue was studied by
\citet{altafini2025multistability}, who also observed three clusters numerically and noted that their stability could
not be verified analytically, and by \citet{altafini2026discrete} for the discrete-time residual update, where the
configurations shown to be stable are aligned with eigenvectors of $V$. So the existence of several attractors is not new in
itself; our contribution is the causal setting, where the first token is pinned to the top eigenvector and the
question is whether later tokens must follow it, with exact stability certificates, exact phase boundaries, and
attractors off every eigenvector. \citet{wu2024role} show that
masks and LayerNorm can prevent collapse in discrete normalized attention, with a two-token causal example in which
the value matrix is a defective two-dimensional shear (a repeated eigenvalue); our examples instead have diagonalizable
value matrices with distinct positive eigenvalues. Other related directions include normalization schemes \citep{karagodin2025normalization},
spectral selection under symmetric couplings \citep{kuehn2026spectral}, Jacobian analyses of recurrent self-attention
\citep{tomihari2025recurrent}, and the special role of the first token in trained models \citep{xiao2024efficient}.

\bibliography{refs}

\appendix

\section{Proof of \texorpdfstring{\cref{prop:full}}{Proposition}}\label{app:full}

By \cref{lem:first}, for every $x_1(0)\notin\{\pm e_2\}$, $x_1(t)\to\sigma e_1$ with $\norm{x_1(t)-\sigma e_1}=O(e^{-\kappa
t})$, $\kappa=\lambda-\mu$. By the symmetry $X\mapsto -X$ it suffices to treat $\sigma=+1$. Let $f$ be the anchored
angular field of $x_2$ for $x_1\equiv e_1$ and write the angle equation of $x_2$ as $\dot\theta=g(t,\theta)$. Since the
right-hand side of \eqref{eq:csa} is smooth on the compact state space, $\norm{g(t,\cdot)-f}_{C^1}=O(e^{-\kappa t})$.

The zeros of $f$ are finitely many (each is hyperbolic, hence isolated, on the compact circle), and around the circle they
alternate between attracting ($f'<0$) and repelling ($f'>0$). Choose disjoint closed arcs $I_k$ around them, so small that
$\abs{f'}$ is bounded below on each, and let $c>0$ bound $\abs f$ from below on the complement. For $t\ge T$ large,
$\norm{g(t,\cdot)-f}_{C^1}<\frac12\min(c,\min_k\min_{I_k}\abs{f'})$; then $g(t,\cdot)$ has the sign of $f$ off the arcs,
points into the arcs around attracting zeros and out of the arcs around repelling zeros. Hence after time $T$ a trajectory
either stays forever in an arc $I_k$ around a repelling zero, or eventually enters an arc around an attracting zero and stays
there. Repeating the construction with smaller arcs shows that in the second case $\theta(t)$ converges to that zero, and in
the first case to the repelling zero.

Fix $x_1(0)$. If two initial angles $\theta(0)\neq\theta'(0)$ both converged to the same repelling zero $\theta^\ast$, then for
$t\ge T'$ both would lie in an arc where $\partial_\theta g\ge c'>0$, and by the mean value theorem their difference would
satisfy $\abs{\theta(t)-\theta'(t)}\ge e^{c'(t-T')}\abs{\theta(T')-\theta'(T')}$, contradicting convergence (the difference is
never $0$ by uniqueness of solutions). So for fixed $x_1(0)$ at most finitely many initial angles converge to repelling zeros,
and by Fubini the set of such initializations is null. Near an attracting zero $\theta^\ast$ with $f'\le-\gamma<0$ on the arc,
$e(t)=\abs{\theta(t)-\theta^\ast}$ satisfies $D^+e\le-\gamma e+Ce^{-\kappa t}$, which gives exponential convergence at every rate
below $\min(\gamma,\kappa)$. \qed

\section{Proofs for \texorpdfstring{\cref{sec:repeated}}{Section 6}}\label{app:repeated}

Throughout, \eqref{eq:H} holds, $\rho_i=\alpha_i=\ip{x_i}{h_i}$, and for a unit $\xi\in L$ we use \eqref{eq:decomp} with
$u_i=c_i=\ip{\xi}{x_i}$, $v_i=z_i=\Pp_Wx_i$ and $y_i=\Pp_{L^\perp}x_i$. Because $V$ acts as $\lambda$ on $L$ and as $B$ on
$L^\perp$,
\begin{equation}\label{eq:components}
\dot u_i=(\lambda p_{ii}-\rho_i)u_i+\lambda\sum_{j<i}p_{ij}u_j,\quad
\dot v_i=(\lambda p_{ii}-\rho_i)v_i+\lambda\sum_{j<i}p_{ij}v_j,\quad
\dot y_i=(p_{ii}B-\rho_iI)y_i+\sum_{j<i}p_{ij}By_j .
\end{equation}
The same equation as for $u_i$ holds for $\ip{h}{x_i}$ for any fixed $h\in L$.

\begin{lemma}[Capture]\label{lem:capture}
Fix $i\ge2$ and a solution with $x_j(t)\to\xi$ for all $j<i$, where $\xi\in L$ is a unit vector. If $u_i(t_0)\ge0$ for some
sufficiently large $t_0$, then $x_i(t)\to\xi$.
\end{lemma}

\begin{proof}
Choose $T$ with $u_j\ge\frac12$ for $j<i$ and $t\ge T$. By \eqref{eq:weightlb},
$H_i:=\lambda\sum_{j<i}p_{ij}u_j\ge h_0:=\lambda(i-1)e^{-2\norm A}/(2i)>0$ for $t\ge T$, so by variation of constants in
\eqref{eq:components}, $u_i(t_0)\ge0$ with $t_0\ge T$ implies $u_i(t)>0$ for all $t>t_0$. For such $t$ put $P_i=v_i/u_i$,
$R_i=y_i/u_i$, $\gamma_{ij}=\lambda p_{ij}u_j/u_i>0$ and $\Gamma_i=\sum_{j<i}\gamma_{ij}=H_i/u_i\ge h_0$. A direct computation
from \eqref{eq:components}, in which the terms with $\rho_i$ cancel, gives
\[
\dot P_i=\sum_{j<i}\gamma_{ij}(P_j-P_i),\qquad
\dot R_i=p_{ii}(B-\lambda I)R_i-\Gamma_iR_i+\lambda^{-1}\sum_{j<i}\gamma_{ij}BR_j .
\]
Taking upper right Dini derivatives of the norms and using $\ip{(B-\lambda I)R}{R}\le-\delta\norm R^2$,
\[
D^+\norm{P_i}\le-\Gamma_i\big(\norm{P_i}-\max_{j<i}\norm{P_j}\big),\qquad
D^+\norm{R_i}\le-\delta p_{ii}\norm{R_i}-\Gamma_i\Big(\norm{R_i}-\frac{\norm B}{\lambda}\max_{j<i}\norm{R_j}\Big).
\]
The predecessor ratios tend to $0$. Given $\varepsilon>0$, for large $t$ the subtracted maxima are below $\varepsilon$; then
$\norm{P_i}$ cannot cross $\varepsilon$ upwards, and while $\norm{P_i}>\varepsilon$ the excess decreases at rate at least
$h_0$. Hence $\limsup\norm{P_i}\le\varepsilon$, and likewise for $R_i$; so $P_i,R_i\to0$. Since
$1=u_i^2(1+\norm{P_i}^2+\norm{R_i}^2)$ and $u_i>0$, $u_i\to1$ and $x_i\to\xi$.
\end{proof}

\begin{proof}[Proof of \cref{thm:fewtokens}]
Normalize $\norm h=1$ and put $w_i=\ip h{x_i}$; it satisfies the $u$-equation of \eqref{eq:components}. Inductively in $i$,
$w_i>0$ for all $t\ge0$: the self-term is linear in $w_i$ and the forcing $\lambda\sum_{j<i}p_{ij}w_j$ is positive. Also
$\ip h\xi>0$, because $\ip h{\Pp_Lx_1(0)}=\ip h{x_1(0)}>0$. Token $1$ converges to $\xi$ by \cref{lem:first}. Suppose all
predecessors of token $i$ converge to $\xi$, and let $r_i=u_i/w_i$. From \eqref{eq:components},
\[
\dot r_i=\sum_{j<i}\frac{\lambda p_{ij}w_j}{w_i}\,(r_j-r_i),
\]
with positive coefficients whose sum is at least $\lambda\sum_{j<i}p_{ij}w_j$ (as $w_i\le1$), which is bounded below for large
$t$ because $w_j\to\ip h\xi>0$. The predecessor ratios converge to $r_\ast=1/\ip h\xi>0$, and the comparison argument of
\cref{lem:capture} applied to $\abs{r_i-r_\ast}$ gives $r_i\to r_\ast$. So $u_i=w_ir_i>0$ eventually, and \cref{lem:capture}
gives $x_i\to\xi$.

If $n\le m$, then for almost every initialization the vectors $\Pp_Lx_1(0),\dots,\Pp_Lx_n(0)$ are linearly independent:
dependence at step $i$ requires $x_i(0)$ to lie in $\operatorname{span}(\Pp_Lx_1(0),\dots,\Pp_Lx_{i-1}(0))\oplus L^\perp$, a
proper subspace when $i\le m$, which meets the sphere in a null set (Fubini). For independent projections there is $h\in L$
with $\ip h{\Pp_Lx_i(0)}=1$ for all $i$, and $\ip h{x_i(0)}=\ip h{\Pp_Lx_i(0)}$.
\end{proof}

\begin{proof}[Proof of \cref{thm:codim1}]
Here $L^\perp=\operatorname{span}(e)$, $B=\mu$, $\delta=\lambda-\mu$, and $W=L\cap\xi^\perp$ has dimension $d-2\ge1$. We show
by induction on $i$ that, for almost every initialization, $x_1,\dots,x_i$ converge to $\xi$; the case $i=1$ is
\cref{lem:first}.

\emph{Step 1: limits exist.} Fix a solution whose first $i-1$ tokens converge to $\xi$. If $u_i\ge0$ at some late time,
\cref{lem:capture} applies. Otherwise $u_i<0$ for all large $t$. With $H=\lambda\sum_{j<i}p_{ij}u_j\ge h_0$ and
$c=\lambda p_{ii}-\rho_i$, $\abs c\le C$ after some time $T$, choose $\varepsilon>0$ with $C\varepsilon\le h_0/2$. On the strip
$-\varepsilon\le u_i\le0$ we have $\dot u_i\ge h_0/2$, so a trajectory entering the strip after $T$ reaches $0$; hence in this
case $u_i\le-\varepsilon$ for all $t\ge T$. Put $s=-u_i\in[\varepsilon,1]$, so $\dot s=cs-H$ and, for $t'\ge t\ge T$,
\[
E(t,t'):=e^{\int_t^{t'}c}=\frac{s(t')}{s(t)}\,e^{\int_t^{t'}H/s}\ge\varepsilon\,e^{h_0(t'-t)} .
\]
With $g=\lambda\sum_{j<i}p_{ij}v_j\to0$, variation of constants and boundedness of $v_i$ give
\[
v_i(t)=-\int_t^\infty E(t,\tau)^{-1}g(\tau)\,d\tau,\qquad
\norm{v_i(t)}\le(\varepsilon h_0)^{-1}\sup_{\tau\ge t}\norm{g(\tau)}\to0 .
\]
Thus $x_i$ approaches the circle in $\operatorname{span}(\xi,e)$, on which we use the angle $\theta=\operatorname{atan2}(\ip
e{x_i},u_i)$, confined to a compact subset of $(\frac\pi2,\frac{3\pi}2)$. With all predecessors at $\xi$ and $v_i=0$, the
token at $\cos\theta\,\xi+\sin\theta\,e$ has total predecessor weight $\omega(\theta)$ and self weight $1-\omega(\theta)$,
and
\[
\dot\theta=f(\theta):=-\sin\theta\,\big(\lambda\omega(\theta)+(\lambda-\mu)(1-\omega(\theta))\cos\theta\big),
\]
a real-analytic function on the circle that is not identically zero ($f(\pi/2)=-\lambda\omega(\pi/2)<0$), hence with finitely many
zeros. Since the predecessors converge and $v_i\to0$, $\dot\theta=f(\theta)+o(1)$. A bounded solution of such an equation
converges to a zero of $f$: if $\liminf\theta<\limsup\theta$, some point strictly between them has $f\ne0$, and near it
$\dot\theta$ eventually has a fixed sign, which forbids crossing it in both directions infinitely often; and the limit must be a
zero. So $x_i\to q^\ast$, an equilibrium of the frozen system, with $\ip{\xi}{q^\ast}<0$.

\emph{Step 2: every bad limit is unstable.} Consider the prefix system $(x_1,\dots,x_i)$ at the equilibrium
$(\xi,\dots,\xi,q^\ast)$. In \eqref{eq:components} the $W$-component of the last token obeys
$\dot v_i=(\lambda p_{ii}-\rho_i)v_i+\lambda\sum_{j<i}p_{ij}v_j$, which vanishes identically when $v_1=\dots=v_i=0$, whatever the
in-plane variables are. Hence the last diagonal block of the Jacobian is block triangular with the eigenvalue
$c^\ast=\lambda(1-\omega)-\rho_i$ on $W$, with $\omega=\omega(\theta^\ast)$. If $q^\ast=-\xi$, then $\rho_i=\lambda(1-2\omega)$ and
$c^\ast=\lambda\omega>0$. If $q^\ast$ is off the axis, its $e$-component is nonzero and its equilibrium equation gives
$\rho_i=(1-\omega)\mu$, so $c^\ast=(1-\omega)(\lambda-\mu)>0$. By
\cref{lem:jac} the prefix Jacobian has the eigenvalue $c^\ast>0$.

\emph{Step 3: a null set.} Let $E$ be the set of all prefix equilibria found in Step~2 (for all unit $\xi\in L$); it need not be
discrete. Let $\varphi$ be the time-one map of the prefix flow. For each $p\in E$, $D\varphi(p)$ has an eigenvalue of modulus
greater than $1$, so by the local center-stable manifold theorem \citep[Thm.~III.7]{shub1987global} there are a neighbourhood
$U_p$ and an embedded disc $W^{cs}_p$ of positive codimension containing every point whose forward $\varphi$-orbit stays in
$U_p$. By second countability, countably many $U_{p_k}$ cover $E$. If a solution converges to some point of $E$, then for some $k$
and some integer $N$ its forward orbit from time $N$ stays in $U_{p_k}$, so its initial point lies in $\varphi^{-N}(W^{cs}_{p_k})$.
The countable union $\bigcup_{k,N}\varphi^{-N}(W^{cs}_{p_k})$ is null because $\varphi$ is a diffeomorphism. Together with the
induction hypothesis (lifted to the $i$-token space, which preserves null sets), this shows that $x_i\to\xi$ for almost every
initialization.
\end{proof}

\begin{proof}[Proof of \cref{prop:rate}]
At consensus $p_{ij}\to1/i$ and $u_i\to1$, so in the proof of \cref{lem:capture} $\Gamma_i\to\lambda(1-\frac1i)\ge\frac\lambda2$
and the $R$-inequality gains the extra damping $\delta p_{ii}$, giving the limiting rate
$(\delta+\lambda(i-1))/i>\min(\delta,\lambda/2)$ for $i\ge2$. For token~1, $P_1=0$ and $R_1=O(e^{-\delta t})$ by \cref{lem:first}.
For large $t$ all damping coefficients exceed some number larger than $\rho$, and inductively in $i$ the integrating-factor
inequality with exponentially decaying input gives $P_i,R_i=O(e^{-\rho t})$; the identity $1=u_i^2(1+\norm{P_i}^2+\norm{R_i}^2)$
converts this into the stated bound. Enlarging $C_\rho$ covers the initial time interval.
\end{proof}

\end{document}